\documentclass[10pt,twocolumn,letterpaper]{article}

\usepackage{cvpr}              % To produce the CAMERA-READY version
\newcommand{\softmax}{\operatorname*{softmax}}
\newcommand{\Var}{\operatorname{Var}}
\newcommand{\minmax}{\operatorname{minmax}}

\newcommand{\Topone}{\text{Top-1}}

\newcommand{\balpha}{\boldsymbol{\alpha}}

\definecolor{cvprblue}{rgb}{0.21,0.49,0.74}
\usepackage[pagebackref,breaklinks,colorlinks,allcolors=cvprblue]{hyperref}
\graphicspath{{figures/}}
\usepackage{algorithm}
\usepackage{algorithmic}
\usepackage{multirow}
\def\paperID{34173} % *** Enter the Paper ID here
\def\confName{CVPR}
\def\confYear{2026}

\title{Label What Matters: Modality-Balanced and Difficulty-Aware \\ Multimodal Active Learning}
 
\author{
	Yuqiao Zeng$^{1}$, Xu Wang$^{1}$, Tengfei Liang$^{1}$, Yiqing Hao$^{1}$, Yi Jin$^{1}$\thanks{Corresponding author.}, Hui Yu$^{2}$\\
	$^{1}$Key Laboratory of Big Data and Artificial Intelligence in Transportation, Ministry of Education;\\
	State Key Laboratory of Advanced Rail Autonomous Operation;\\
	School of Computer and Information Technology, Beijing Jiaotong University, Beijing, China\\
	$^{2}$School of Psychology \& Neuroscience, University of Glasgow, Glasgow, UK\\
	{\small\ttfamily \{yuqiaozeng, xu.wang, tengfei.liang, yiqinghao, yjin\}@bjtu.edu.cn}\\
	{\small\ttfamily Hui.Yu@glasgow.ac.uk}
}
\begin{document}
	\maketitle
	
\begin{abstract}
	Multimodal learning integrates complementary information from different modalities such as image, text, and audio to improve model performance, but its success relies on large-scale labeled data, which is costly to obtain. Active learning (AL) mitigates this challenge by selectively annotating informative samples. In multimodal settings, many approaches implicitly assume that modality importance is stable across rounds and keep selection rules fixed at the fusion stage, which leaves them insensitive to the dynamic nature of multimodal learning, where the relative value of modalities and the difficulty of instances shift as training proceeds. To address this issue, we propose \textbf{RL-MBA}, a reinforcement-learning framework for modality-balanced, difficulty-aware multimodal active learning. RL-MBA models sample selection as a Markov Decision Process, where the policy adapts to modality contributions, uncertainty, and diversity, and the reward encourages accuracy gains and balance. Two key components drive this adaptability: (1) Adaptive Modality Contribution Balancing (AMCB), which dynamically adjusts modality weights via reinforcement feedback, and (2) Evidential Fusion for Difficulty-Aware Policy Adjustment (EFDA), which estimates sample difficulty via uncertainty-based evidential fusion to prioritize informative samples. Experiments on Food101, KineticsSound, and VGGSound demonstrate that RL-MBA consistently outperforms strong baselines, improving both classification accuracy and modality fairness under limited labeling budgets.
\end{abstract}

\begin{figure}[!t]
	\centering
	\includegraphics[width=0.9\linewidth]{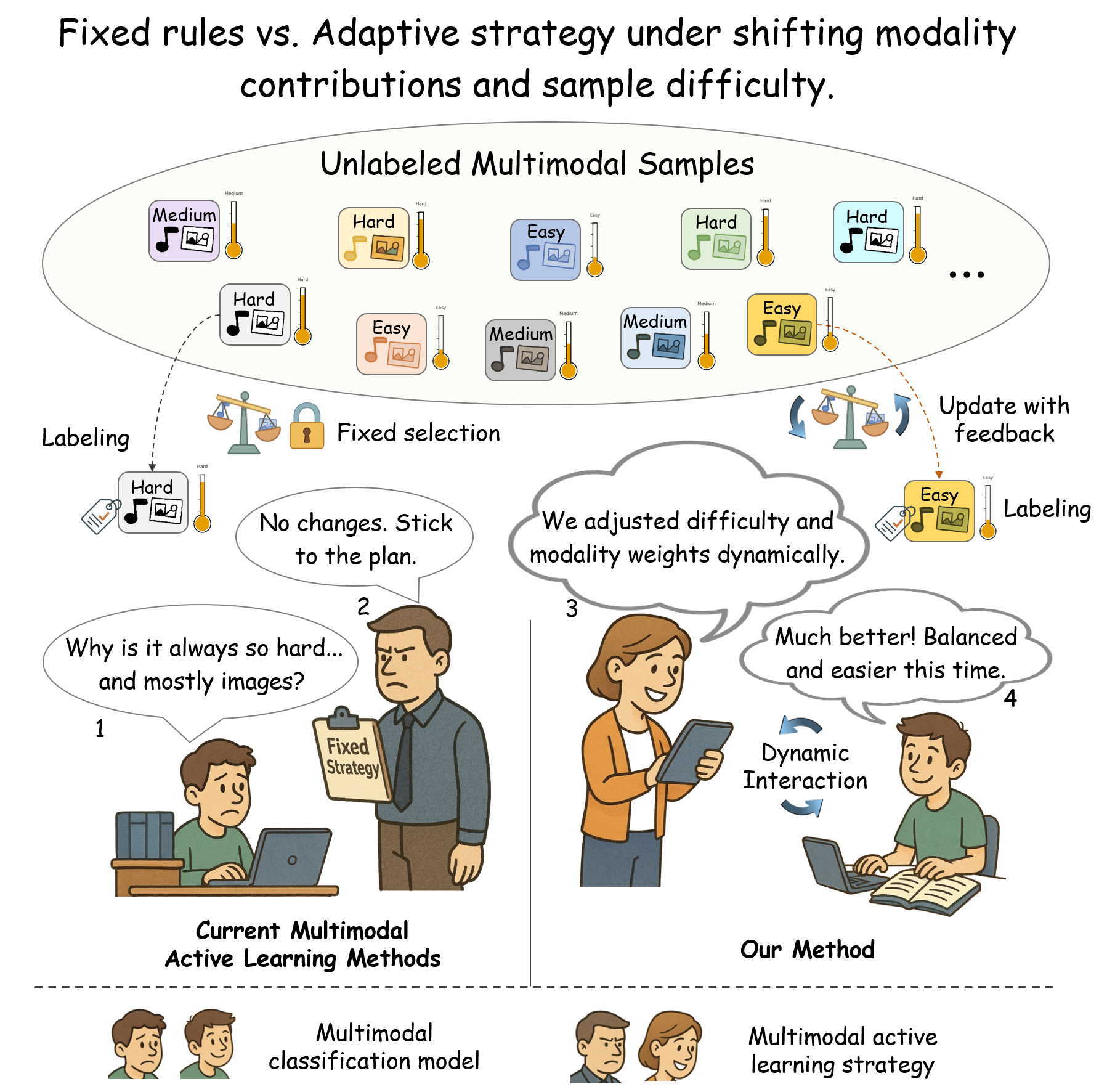}
	\caption{\textbf{Fixed rules vs.\ adaptive strategy in multimodal active learning.}
		The unlabeled pool contains items with different difficulty levels and different dominant modality.
		\emph{Left} (1–2): fixed strategies keep selecting hard, dominant-modality samples and do not adjust when the value of each modality or the difficulty of samples changes across rounds.
		\emph{Right} (3–4): an adaptive strategy reweights modality contributions and difficulty, yielding more balanced, budget-efficient batches.}
	\label{challenge}
\end{figure}

\section{Introduction}

Multimodal learning has attracted growing attention because it combines information from different sources, enabling models to build richer representations and achieve stronger performance than single-modal approaches in different tasks \cite{ash2019deep, baltruvsaitis2018multimodal,zeng2024mmi}. Despite these benefits \cite{guo2019deep}, collecting such annotations, especially across multiple modalities, is costly and time-consuming, creating a major barrier for deploying multimodal models in practice. Active learning (AL) is therefore used to lower labeling costs by selectively querying the most informative samples for annotation, reducing the need for large amounts of labeled data \cite{settles2011theories,parvaneh2022active,schmidt2025joint}.

However, most active learning methods in multimodal settings still rely on fixed selection rules, which fail to address the dynamic nature of multimodal learning \cite{shen2023towards,caramalau2021sequential,sener2017active,ducoffe2018adversarial,gal2017deep,ash2019deep}. These methods often suffer from a \textit{modality imbalance}: batches are biased toward samples where a strong modality dominates, while weaker modalities are underused. This imbalance reduces cross-modal complementarity and hinders generalization \cite{beluch2018power,margatina2021active}. As illustrated in Fig.~\ref{challenge} (left), fixed strategies fail to react when the value of each modality or the difficulty of samples shifts from one round to the next, making the budget less effective and limiting fusion quality. Existing methods which attempt to reduce this bias, such as BMMAL \cite{shen2023towards}, rely on static adjustments during training; they help to some extent but implicitly assume that modality importance is stable across rounds, which rarely holds as the model and labeled pool evolve. These observations point to a simple requirement: \emph{sampling should update from feedback}. Concretely, a good multimodal AL strategy should (i) reweight modality contributions across rounds, so that rising modalities are used and fading ones do not dominate, and (ii) use quantitative uncertainty to focus on challenging but informative cases while avoiding unproductive extremes. Such a rule aims for balanced, long-term gains under a fixed budget rather than a fixed strategy. This highlights the need for a more adaptive approach, where the sampling strategy evolves based on feedback from each learning round. Such a dynamic strategy would enable the model to continually adjust its focus on modalities and sample difficulty as training progresses, addressing the limitations of static selection rules.

To this end, we propose \emph{RL-MBA}, a feedback-driven multimodal AL method that makes the sampling rule adaptive along these two axes. RL-MBA treats sample selection as a policy that is updated from round-level feedback (accuracy gains and balance signals), where a unified score first proposes a compact candidate set and the policy selects the final query batch from these candidates. A natural way to realize such feedback-driven updates is to model selection as a Markov Decision Process (MDP) with a policy optimized for long-term reward; this allows the policy to adjust to the current model state, the distribution of the unlabeled pool, and the changing value of each modality over time. In Fig.~\ref{challenge} (right), the adaptive policy yields balanced batches by reweighting modalities and by prioritizing samples whose difficulty is informative for learning.

RL-MBA introduces two components to implement this adaptivity:
(i) \emph{Adaptive Modality Contribution Balancing (AMCB):} estimates modality contributions on a fixed validation split and reweights them across rounds, coupling the same weights with fusion, scoring, and the policy state.
(ii) \emph{Evidential Fusion for Difficulty-Aware Adjustment (EFDA):} computes calibrated, evidential uncertainty and uses it to emphasize challenging but useful samples, rather than merely the hardest ones. Our main contributions are as follow:
\begin{itemize}
	\item We present RL-MBA, a multimodal active learning framework that updates its sampling rule from feedback, achieving adaptive balance across modalities and difficulty over rounds.
	\item We design \emph{AMCB} to dynamically reweight modality contributions and feed the same weights into fusion, scoring, and policy state for coherent emphasis.
	\item We introduce \emph{EFDA}, an evidential uncertainty module that guides selection toward challenging yet informative samples with improved calibration.
\end{itemize}

\section{Related Work}

\subsection{Active Learning}
Active learning (AL) aims to improve model performance by selectively annotating the most informative unlabeled samples, reducing labeling costs. Existing strategies generally fall into three categories: uncertainty-based, diversity-based, and hybrid approaches. Uncertainty-based methods measure model confidence using metrics such as entropy or prediction margins \cite{settles2011theories, roth2006margin}, with improvements via Monte Carlo dropout for Bayesian uncertainty \cite{gal2016dropout} and multi-model variance estimation \cite{beluch2018power}. Diversity-based methods, such as CoreSet \cite{sener2017active} and determinantal point processes \cite{biyik2019batch}, aim for broad representation but often overlook sample informativeness. Hybrid strategies like BADGE \cite{ash2019deep} combine uncertainty with diversity using gradient embeddings, yet their adaptability to multimodal data and dynamic modality importance remains limited, motivating more integrated and adaptive solutions.

\subsection{Reinforcement Learning}
Reinforcement learning (RL) has been explored as an adaptive alternative for active learning, enabling policies to evolve through interaction and reward optimization. Early efforts include RAL \cite{haussmann2019deep}, which uses uncertainty-aware policies, and DRAL \cite{liu2019deep}, which integrates feedback for person re-identification. Policy-based Active Learning \cite{fang2017learning} and LSTM-based agents for few-shot learning \cite{woodward2017active} demonstrate RL’s potential to surpass static heuristics. However, most RL-driven approaches focus on unimodal scenarios and lack mechanisms to address multimodal imbalance or incorporate difficulty-aware selection. Our RL-MBA framework addresses these gaps by jointly modeling modality balance and sample difficulty within a unified RL-based design.

\section{Method}
\label{sec:method}

\begin{figure*}[!t]
	\centering
	\includegraphics[width=17.5cm]{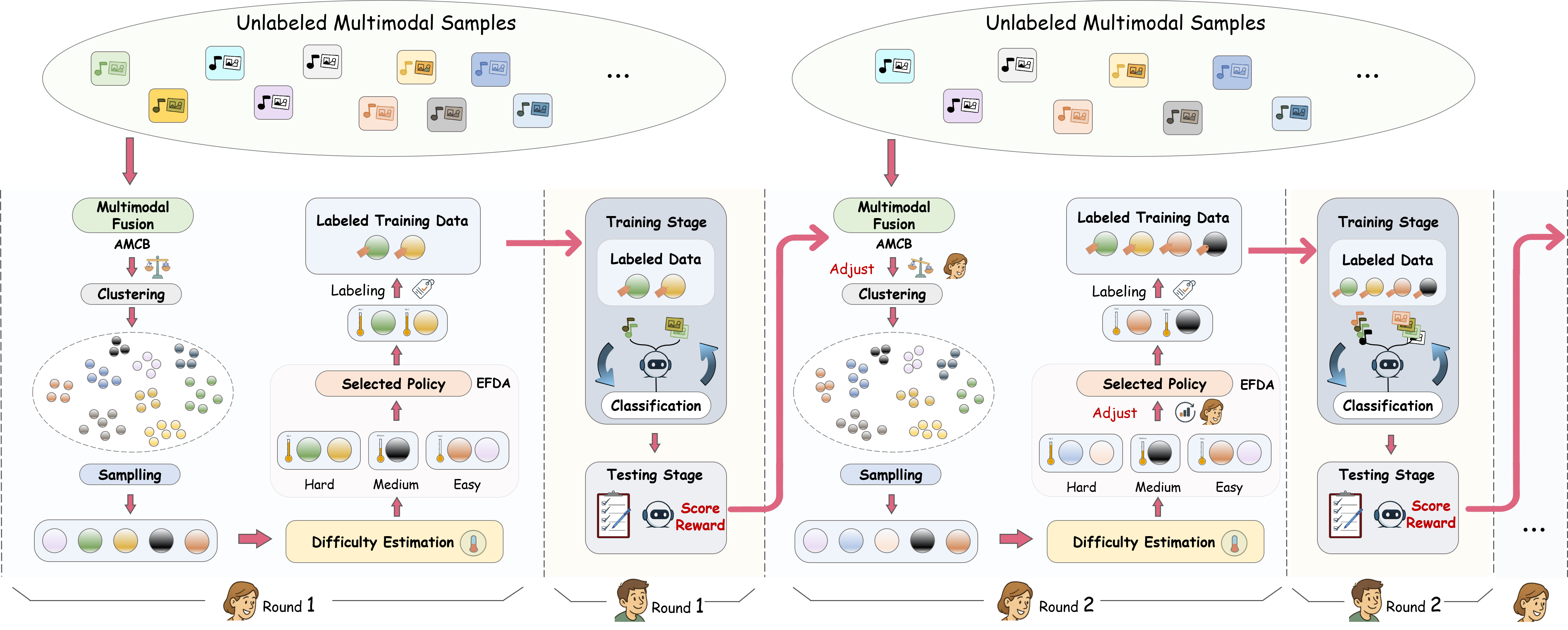}
	\caption{Overview of RL-MBA. Each round consists of (1) multimodal fusion and clustering, (2) evidential uncertainty and difficulty estimation, (3) scoring to form a candidate set, (4) policy-based set selection, (5) retraining, and (6) reward and policy update.}
	\label{fig:framework}
\end{figure*}

\subsection{Problem Definition}
\label{sec:problem}
We consider a multimodal active learning scenario with dataset $\mathcal{D}=\{(\mathbf{x}_i,y_i)\}_{i=1}^N$, where $\mathbf{x}_i=(\mathbf{x}_i^{(1)},\dots,\mathbf{x}_i^{(M)})$ spans $M$ modalities and $y_i\in\{1,\dots,C\}$.
Initially, a small labeled subset $\mathcal{L}\subset\mathcal{D}$ and a large unlabeled pool $\mathcal{U}=\mathcal{D}\setminus\mathcal{L}$ are given.
At each round $t$, a batch of size $b$ is selected from $\mathcal{U}_t$ for annotation under a fixed labeling budget.
We denote the fused multimodal feature by $f(x)$, the modality weights by $\boldsymbol{w}=(w_1,\dots,w_M)$ with $w_m\!\in\![0,1],\sum_m w_m\!=\!1$, and the Dirichlet evidence from each modality by $\balpha_m\!\in\!\mathbb{R}^C_{>0}$.

\subsection{Overview of RL-MBA}
\label{sec:overview}
RL-MBA formulates multimodal sample selection as a Markov Decision Process (MDP) and optimizes a lightweight selection policy via policy-gradient reinforcement learning.
Each active learning round (Fig.~\ref{fig:framework}) proceeds as:
\textbf{(1)} multimodal fusion with adaptive modality weighting (AMCB, Sec.~\ref{sec:amcb}) and budgeted clustering for diversity;
\textbf{(2)} evidential uncertainty and difficulty estimation (EFDA, Sec.~\ref{sec:efda});
\textbf{(3)} unified scoring to construct a candidate set (Sec.~\ref{sec:scoring});
\textbf{(4)} policy-based set selection from the candidates (Sec.~\ref{sec:mdp}, Sec.~\ref{sec:policy});
\textbf{(5)} retraining on the updated labeled set $\mathcal{L}$; and
\textbf{(6)} reward computation and policy update.
\emph{All evaluations use a fixed stratified validation split across rounds.} Per-modality heads share the backbone; at each round they are evaluated on the validation split to compute modality contributions and calibration statistics.

\begin{figure}[!t]
	\centering
	\includegraphics[width=1\columnwidth]{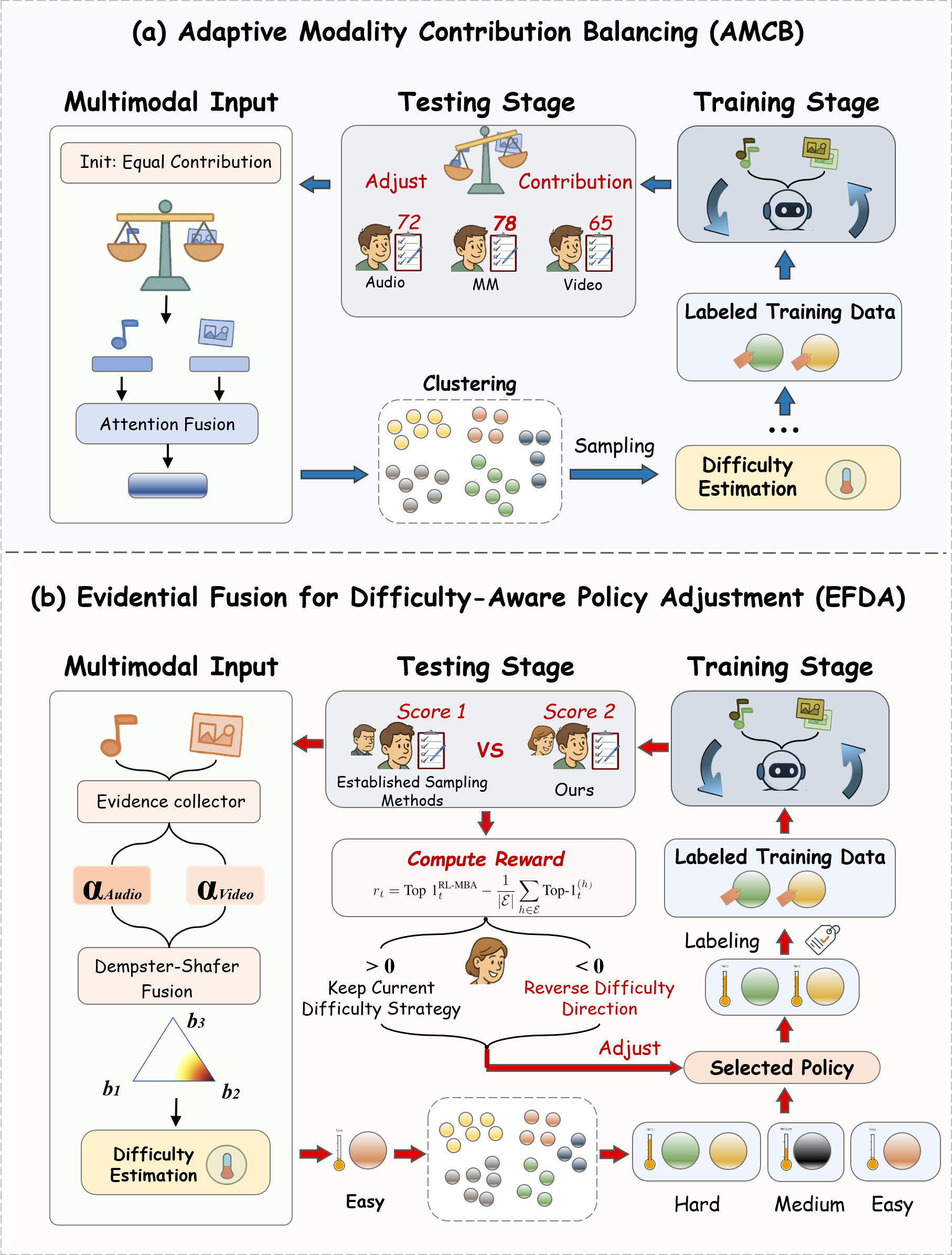}
	\caption{Core components. (a) AMCB updates simplex weights $\boldsymbol{w}$ from round-level modality gains. (b) EFDA fuses Dirichlet evidence to obtain calibrated uncertainty/difficulty; the RL reward updates the selection policy (and optionally uncertainty heads if enabled in implementation).}
	\label{fig:inno}
\end{figure}

\subsection{Adaptive Modality Contribution Balancing (AMCB)}
\label{sec:amcb}
\noindent\textbf{Motivation.} In multimodal AL, a modality’s contribution depends on the current training context: which categories are prevalent or missing, how many labels are available, and how well the model already fits the data. It changes across rounds. Fixed weights bias selection toward the channel that happens to dominate and leave other cues underused. AMCB replaces fixed weighting with a feedback-updated simplex of modality contributions. The same weights guide fusion, scoring, and (part of) the state given to the policy, so emphasis shifts coherently as the task evolves, yielding more balanced and budget-efficient batches.

\noindent\textbf{Design.}
We quantify the current contribution of modality $m$ by the Top-1 gap:
\begin{equation}
	\label{eq:top1-gap}
	\Delta_m=\Topone_m-\Topone_{\text{mm}},
\end{equation}
computed on the fixed validation split at each round (per-modality heads share the backbone and are evaluated with their respective branches).
A positive gap indicates complementary signal beyond the multimodal head; negative gaps indicate redundancy or noise.
Contributions are mapped to a probability simplex by a temperature-controlled softmax
\begin{equation}
	\label{eq:amcb}
	\boldsymbol{w}=\softmax(\boldsymbol{\Delta}/\tau),\qquad w_m\in[0,1],\;\sum_{m=1}^{M} w_m=1,
\end{equation}
where $\tau>0$ modulates sensitivity.
To avoid vanishing modalities we use an optional floor $\varepsilon>0$:
$\tilde w_m=(w_m+\varepsilon)/\sum_j(w_j+\varepsilon)$.

\noindent\textbf{Properties.}
(i) \emph{Simplex consistency.} $\boldsymbol{w}$ is non-negative and sums to one, so fusion is a convex combination and scale-stable.  
(ii) \emph{Monotonicity and control.} $\partial w_m/\partial \Delta_m \propto (1/\tau)\,w_m(1-w_m)$; smaller $\tau$ yields quicker emphasis shifts when a modality becomes informative.  
(iii) \emph{Identity and robustness.} If a single modality dominates ($\Delta_k\!\gg\!\Delta_{m\neq k}$), then $\boldsymbol{w}\!\to\!e_k$; if contributions tie, $\boldsymbol{w}$ is uniform, preventing premature specialization.  

\noindent\textbf{Coupling across the pipeline.}
The same $\boldsymbol{w}$ is injected into three loci:
(i) \emph{Fusion:} $f(x)=\sum_m w_m f_m(x)$ focuses representation on currently informative views;
(ii) \emph{Scoring:} $q(x)$ (Sec.~\ref{sec:scoring}) weighs modality-wise uncertainties by $\boldsymbol{w}$, aligning acquisition with contribution;
(iii) \emph{State:} the state includes $\boldsymbol{w}$ (or $\boldsymbol{\Delta}$), exposing the balance to the policy so that future selection can react to shifts in modality value.

\subsection{Evidential Fusion for Difficulty-Aware Estimation (EFDA)}
\label{sec:efda}
\noindent\textbf{Motivation.}
Uncertainty should reflect both aleatoric variability (intrinsic noise) and epistemic lack of evidence.
Combining signals at the posterior level (e.g., simple products or averages) may become overconfident when calibration differs across modalities or when a local failure occurs.
EFDA instead performs \emph{evidence}-level fusion that is additive, bounded, and aligned with AMCB: modalities with higher $w_m$ contribute more evidence, while weaker modalities neither dominate nor collapse the estimate.

\noindent\textbf{Design.}
Each modality head outputs Dirichlet evidence $\balpha_m(x)\in\mathbb{R}_{>0}^C$.
We fuse evidence additively,
\begin{equation}
	\label{eq:efda-fuse}
	\balpha_f(x)=\mathbf{1}+\sum_{m=1}^{M} w_m\big(\balpha_m(x)-\mathbf{1}\big),
\end{equation}
so that the fused prior $\balpha_f$ interpolates between modalities according to $\boldsymbol{w}$.
Per-modality logits can be temperature-scaled on the validation split (one scalar per modality) before converting to evidence, yielding better-calibrated \emph{per-modality} and \emph{fused} Dirichlet distributions.

\noindent\textbf{Rationale.}
Additive evidence preserves identity when a single modality is trusted ($\boldsymbol{w}=e_k$ gives $\balpha_f=\balpha_k$) and keeps confidence within explicit bounds ($1\le \alpha_{f,c}\le 1+\sum_m w_m(\alpha_{m,c}-1)$), preventing runaway certainty.
It degrades gracefully under weak or missing inputs because small $w_m$ or low evidence limits a modality’s influence, and it remains calibration-friendly since scaling is applied upstream of evidence.

\noindent\textbf{Difficulty-aware uncertainty.}
Given $\balpha_f$, the Dirichlet predictive variance
\begin{equation}
	\label{eq:efda-unc}
	\Var[p_c]=\frac{\alpha_{f,c}(\alpha_{f,0}-\alpha_{f,c})}{\alpha_{f,0}^2(\alpha_{f,0}+1)},\qquad
	\mathsf{U}(x)=\tfrac{1}{C}\sum_{c=1}^C \Var[p_c],
\end{equation}
with $\alpha_{f,0}=\sum_c \alpha_{f,c}$, serves as a \emph{difficulty} proxy: samples with diffuse posteriors (small $\alpha_{f,0}$ or balanced class mass) receive larger $\mathsf{U}(x)$ and are prioritized.
This naturally couples with AMCB: if a modality becomes more informative, $w_m$ increases, its evidence contributes more to $\balpha_f$, and the uncertainty of easy samples shrinks, freeing budget for genuinely difficult cases.

\subsection{Scoring with AMCB, EFDA and Diversity}
\label{sec:scoring}
Let $u_m(x)$ denote modality-wise uncertainty derived analogously to Eq.~\ref{eq:efda-unc} on $\balpha_m$, and $f(x)=\sum_m w_m f_m(x)$ the fused feature.
To promote representativeness, we apply budgeted $k$-means++ on $\{f(x)\,|\,x\in\mathcal{U}_t\}$ with $k{=}b$ centroids $\{c_k\}_{k=1}^b$ (at most 5 iterations).
The nearest-centroid distance is
\begin{equation}
	\label{eq:div}
	d(x)=\min_{k\le b}\,\lVert f(x)-c_k\rVert_2,
\end{equation}
which encourages coverage of underrepresented regions rather than selecting outliers.
Within each round, uncertainties and distances are min–max normalized to mitigate scale drift:
$\tilde u_m(x)=\minmax(u_m(x))$, $\tilde d(x)=\minmax(d(x))$.
The final score combines informativeness and diversity:
\begin{equation}
	\label{eq:score}
	q(x)=\sum_{m=1}^M w_m\,\tilde u_m(x)+\beta\,\tilde d(x).
\end{equation}
We use $q(\cdot)$ to build a candidate set $C_t$ (Sec.~\ref{sec:mdp}) rather than directly taking $\mathrm{Top}\mbox{-}b$ as the action.

\subsection{MDP Formulation with State, Action and Reward}
\label{sec:mdp}
\noindent\textbf{State $s_t$.}
A fixed-length vector $s_t=[g_t \,\|\, \phi_t \,\|\, \bar{u}_t \,\|\, \bar{d}_t \,\|\, \rho_t]$, including
(i) validation statistics $g_t$ (\Topone, NLL, ECE; all normalized to $[0,1]$);
(ii) modality contributions $\phi_t$ (Top-1 gaps);
(iii) aggregated uncertainty and diversity summaries $(\bar{u}_t,\bar{d}_t)$; and
(iv) training diagnostics $\rho_t$ (loss slope, gradient norm).
All elements are standardized across rounds.

\noindent\textbf{Action $a_t$.}
We first construct a candidate set
\begin{equation}
	\label{eq:cand}
	C_t=\mathrm{Top}\mbox{-}K\ \text{by}\ q(\cdot)\ \text{on}\ \mathcal{U}_t,\qquad K=\kappa b,
\end{equation}
where $\kappa\!>\!1$ is a small constant (e.g., $\kappa\!=\!5$ in our implementation).
The action is a \emph{set} of $b$ samples selected from $C_t$ \emph{without replacement} by the policy $\pi_\theta(\cdot|s_t)$ (Sec.~\ref{sec:policy}).

\noindent\textbf{Transition.}
Each step involves candidate construction, policy-based selection, labeling, retraining for $E$ epochs, and recomputation of validation statistics,
resulting in a deterministic MDP given the dataset and fixed training protocol (up to stochastic optimization noise).

\noindent\textbf{Reward $r_t$.}
We use a round-level reward based on the validation \Topone{} of the multimodal model, optionally defined \emph{relative} to precomputed baseline learning curves:
\begin{align}
	\label{eq:reward}
	r_t &= \Topone^{\text{RL-MBA}}_t
	- \frac{1}{|\mathcal{E}|}\sum_{h\in\mathcal{E}}
	\Topone^{(h)}_t, \\
	\mathcal{E} &= \{\textsc{GCNAL},\textsc{Badge},\textsc{Bmmal}\}.
\end{align}
Importantly, the baseline scores $\{\Topone^{(h)}_t\}$ are \emph{precomputed offline} under the same protocol and read as constants during RL-MBA training; hence, computing $r_t$ does not require concurrent training of baselines and adds negligible overhead.
In practice, we further smooth the reward with an exponential moving average to improve stability.

\noindent\textbf{Per-round complexity.}
Let $F$ be the backbone forward cost per sample and $M$ the number of modalities.
Each round requires $O(|\mathcal{U}_t|F)$ for feature extraction, $O(|\mathcal{U}_t|M)$ for scoring (AMCB/EFDA are linear in $M$), and $O(|\mathcal{U}_t|\log|\mathcal{U}_t|)$ for sorting; budgeted $k$-means++ uses $k{=}b$ with a small fixed number of iterations.
Candidate construction adds only sorting-based overhead, while policy selection operates on $K=\kappa b$ candidates and is negligible compared to retraining.

\subsection{Policy Network and Optimization}
\label{sec:policy}
\noindent\textbf{Policy parameterization.}
The policy is a lightweight MLP that, given the round state $s_t$, outputs selection logits for candidates in $C_t$.
We sample a batch of size $b$ sequentially \emph{without replacement}. Let $C_t^{(j)}$ denote the remaining candidates at step $j$ ($C_t^{(1)}=C_t$).
At each step $j$, the policy defines
\begin{equation}
	\label{eq:step-prob}
	\pi_\theta(x\,|\,s_t, C_t^{(j)})=
	\frac{\exp(z_\theta(s_t,x))}{\sum_{x'\in C_t^{(j)}} \exp(z_\theta(s_t,x'))},
\end{equation}
and samples $x_{i_j}\sim \pi_\theta(\cdot|s_t, C_t^{(j)})$, then removes it from the candidate set.
Thus, the probability of the batch action $a_t=\{x_{i_1},\dots,x_{i_b}\}$ is
\begin{equation}
	\label{eq:set-prob}
	\pi_\theta(a_t|s_t)=\prod_{j=1}^{b}\pi_\theta\!\big(x_{i_j}\,\big|\,s_t, C_t^{(j)}\big),
\end{equation}
\begin{equation}
	\label{eq:set-logprob}
	\log\pi_\theta(a_t|s_t)=\sum_{j=1}^{b}\log\pi_\theta\!\big(x_{i_j}\,\big|\,s_t, C_t^{(j)}\big).
\end{equation}
\noindent\textbf{Policy optimization.}
We optimize $\theta$ with REINFORCE using a one-step return at each AL round:
\begin{equation}
	\label{eq:reinforce}
	\nabla_{\theta} J(\theta)=
	\mathbb{E}\!\left[\sum_{t=1}^{T} A_t\,\nabla_{\theta}\log \pi_{\theta}(a_t\mid s_t)\right],
	\qquad
	A_t = r_t - b_t,
\end{equation}
where $b_t$ is a moving-average baseline of past rewards (variance reduction). In implementation we optionally clip $A_t$ to stabilize training.
This formulation lets the policy adapt to the evolving data distribution while keeping the RL component lightweight.
We train one policy per dataset/task under the same AL protocol.

\subsection{Overall Algorithm}
\label{sec:algo}
\noindent
Algorithm~\ref{alg:RL-MBA} summarizes one active learning episode of RL\textendash MBA.
Given a labeled set $\mathcal{L}$, an unlabeled pool $\mathcal{U}$, and a labeling budget $b$ per round, the agent observes the state $s_t$, scores the unlabeled pool to form a candidate set $C_t$, samples an action batch $a_t$ from $C_t$, updates the backbone on $\mathcal{L}$, and receives a (relative) Top-1 reward (Eq.~\ref{eq:reward}).
AMCB (Eq.~\ref{eq:amcb}) adapts modality weights across rounds, while EFDA provides calibrated uncertainty for difficulty-aware scoring.
Per-round cost follows Sec.~\ref{sec:mdp} and the policy update uses Eq.~\ref{eq:reinforce}.

\begin{algorithm}[t]
	\caption{\textbf{RL-MBA:} Modality-Balanced Multimodal Active Learning.
		\emph{Inputs:} dataset $\mathcal{D}=(\mathcal{L},\mathcal{U})$, batch $b$, rounds $T$, epochs $E$, temperature $\tau$, diversity weight $\beta$, candidate multiplier $\kappa$, discount $\gamma$.
		\emph{Output:} trained backbone and policy $\pi_\theta$.}
	\label{alg:RL-MBA}
	\begin{algorithmic}[1]
		\STATE Initialize policy $\theta$; fix a stratified validation split
		\STATE Estimate initial contributions $\Delta$ on the validation split; set $w \leftarrow \mathrm{softmax}(\Delta/\tau)$ \hfill (Eq.~\ref{eq:amcb})
		\FOR{$t=1,\dots,T$}
		\STATE Build state $s_t=[g_t\,\|\,\phi_t\,\|\,\bar u_t\,\|\,\bar d_t\,\|\,\rho_t]$ \hfill (Sec.~\ref{sec:mdp})
		\STATE Run budgeted $k$-means++ on fused features ($k{=}b$, $\le$5 iters); obtain centroids $\{c_k\}_{k=1}^b$
		\FOR{each $x\in\mathcal{U}_t$}
		\STATE Obtain per-modality evidence $\alpha_m(x)$; fuse $\alpha_f(x)=\mathbf{1}+\sum_m w_m(\alpha_m(x)-\mathbf{1})$ \hfill (Eq.~\ref{eq:efda-fuse})
		\STATE Compute $u_m(x)$ via Dirichlet variance; set $d(x)=\min_{k\le b}\|f(x)-c_k\|_2$ \hfill (Eq.~\ref{eq:efda-unc}, Eq.~\ref{eq:div})
		\STATE Normalize: $\tilde u_m(x)=\mathrm{minmax}(u_m(x))$, $\tilde d(x)=\mathrm{minmax}(d(x))$
		\STATE Score: $q(x)=\sum_m w_m\,\tilde u_m(x)+\beta\,\tilde d(x)$ \hfill (Eq.~\ref{eq:score})
		\ENDFOR
		\STATE Construct candidate set $C_t \leftarrow \mathrm{Top}\mbox{-}K$ by $q(\cdot)$ on $\mathcal{U}_t$, $K=\kappa b$ \hfill (Eq.~\ref{eq:cand})
		\STATE Sample action batch $a_t \sim \pi_\theta(\cdot|s_t)$ from $C_t$ without replacement \hfill (Eq.~\ref{eq:set-prob})
		\STATE Query labels $y(a_t)$; $\mathcal{L}\leftarrow \mathcal{L}\cup a_t,\quad \mathcal{U}\leftarrow \mathcal{U}\setminus a_t$
		\STATE Retrain backbone for $E$ epochs
		\STATE Compute reward $r_t$ (Eq.~\ref{eq:reward}); update policy with REINFORCE (Eq.~\ref{eq:reinforce})
		\STATE Recompute $\Delta$ on the validation split; update $w \leftarrow \mathrm{softmax}(\Delta/\tau)$ \hfill (Eq.~\ref{eq:amcb})
		\ENDFOR
	\end{algorithmic}
\end{algorithm}

\section{Experiment}

We evaluate RL-MBA on multiple multimodal datasets to test its ability to rebalance modality contributions and prioritize informative samples. We describe datasets, setup, baselines, and metrics, then present modality-contribution analysis and ablations, combining quantitative and visual evidence of RL-MBA’s advantages.

\subsection{Datasets}
We evaluate RL-MBA on three multimodal datasets that cover distinct modality combinations and semantic complexities.
\textbf{Food101} contains 101 food categories with paired image and text modalities, including 45,719 training and 15,294 test samples.
Each class is associated with a concise textual description that complements the visual modality\cite{zhang2017learning}. \textbf{KineticsSound} comprises 31 action classes combining video and audio modalities, with 14,739 training and 2,594 test clips. \textbf{VGGSound} extends this to 309 diverse sound–vision categories, including 105,243 training and 7,109 test samples.

\begin{figure}[t]
	\centering
	\includegraphics[width=8.5cm]{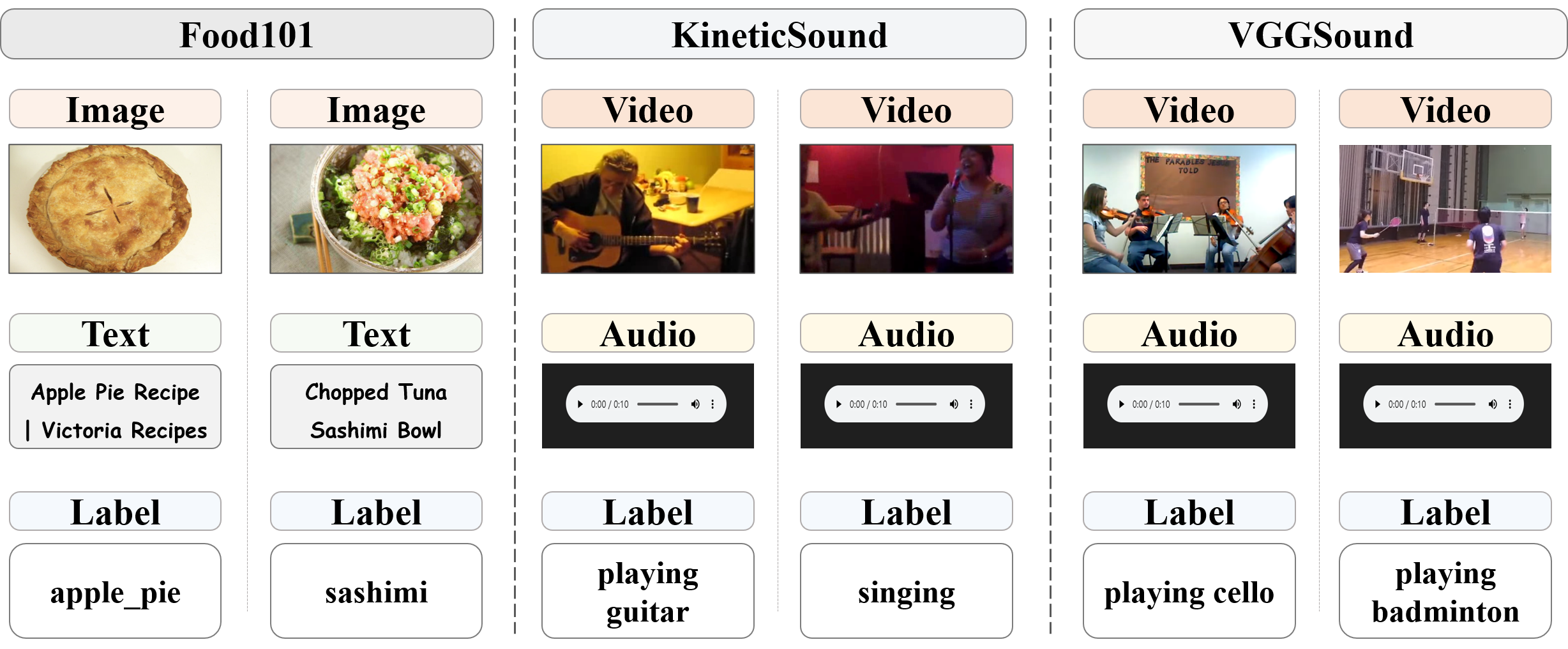}
	\caption{The example from Food101, KineticsSound and VGGSound}
	\label{fig:exp_dataset}
\end{figure}

\subsection{Experimental Setup}
Our experiments are conducted on a system with an NVIDIA RTX 5880 GPU and 128GB of memory, running Ubuntu 18.04.6 LTS as the operating system. We utilize PyTorch version 2.5.1 with CUDA 12.4 support for deep learning tasks, ensuring compatibility with the GPU hardware. We evaluate RL-MBA on two multimodal classification tasks: image-text (Food101) and video-audio (VGGSound, KineticsSound), using task-specific architectures. For image-text classification, we use ResNet-101 (ImageNet-pretrained) for images and BERT-base \cite{devlin2019bert} for text, followed by a shared fully connected layer. Models are optimized with AdamW \cite{loshchilov2017decoupled} for 15 epochs per active learning cycle. Standard image augmentations (random crop, horizontal flip, grayscale) are applied. For video-audio classification, we adopt ResNet2P1D-18 \cite{tran2018closer} for video (pretrained on Kinetics-400 for VGGSound, random init for KineticsSound) and a modified ResNet-18 for audio. Videos are sampled at 10 FPS; audio is converted to spectrograms using a 512-sample window with 353-sample overlap.

\begin{table}[t]
	\centering
	\fontsize{10}{12}\selectfont
	\setlength{\tabcolsep}{1.2mm}
	\begin{tabular}{lccc}
		\toprule
		\textbf{Method} & \textbf{Food101} & \textbf{KineticsSound} & \textbf{VGGSound} \\
		\midrule
		Random   & 0.8470 & 0.4650 & 0.2173 \\
		Entropy  & 0.8480 & 0.4650 & 0.2043 \\
		GCNAL    & 0.8510 & 0.4600 & 0.2033 \\
		CoreSet  & 0.8422 & 0.4600 & 0.2013 \\
		DeepFool & 0.8500 & 0.4680 & 0.1973 \\
		BALD     & 0.8450 & 0.4550 & 0.1993 \\
		BADGE    & 0.8420 & 0.4700 & 0.2023 \\
		BMMAL    & 0.8609 & 0.4745 & 0.2053 \\
		\textbf{RL-MBA} & \textbf{0.8650} & \textbf{0.4841} & \textbf{0.2223} \\
		\bottomrule
	\end{tabular}
	\caption{Top-1 Accuracy on multimodal datasets with \textbf{3,000 labelled samples}. 
		This corresponds to \textbf{6.6\%} of Food101, \textbf{20.4\%} of KineticsSound, and \textbf{2.9\%} of VGGSound training sets.}
	\label{tab:top1_results}
\end{table}

\begin{figure*}[t]
	\centering
	\includegraphics[width=16.5cm]{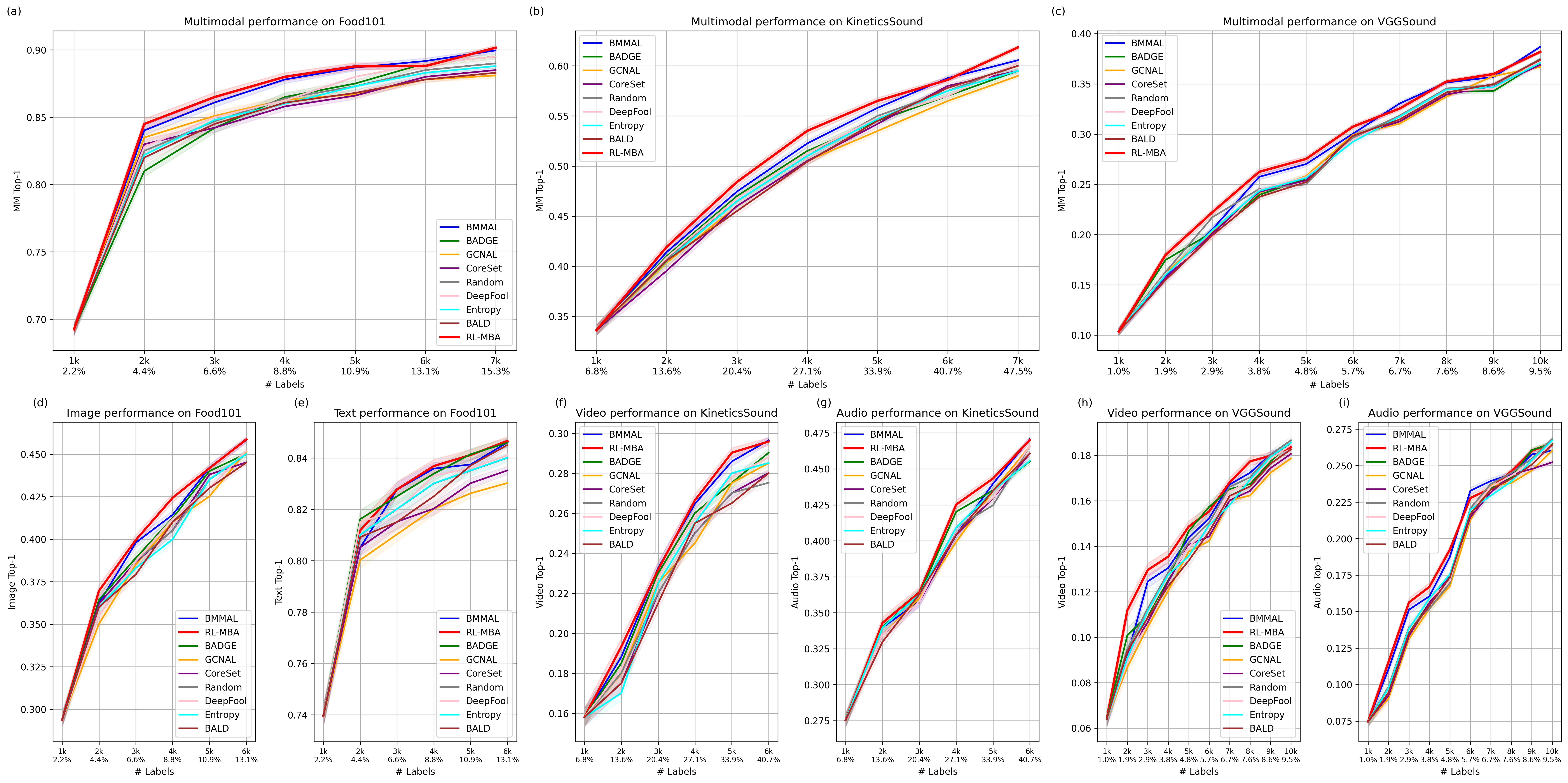}
	\caption{Performance comparison between the proposed method and other conventional AL strategies on Food101, KineticsSound, and VGGSound. The metric selected is top-1 accuracy (Top-1) on multimodal and unimodal classification.}
	\label{fig:performance_curve}
\end{figure*}

\subsection{Baselines}
We compare RL-MBA with widely used active learning strategies. \textbf{Random} selects samples uniformly as a naive baseline. \textbf{Entropy} \cite{settles2011theories} ranks samples by prediction uncertainty. \textbf{GCNAL} \cite{caramalau2021sequential} uses graph convolutional networks for uncertainty and diversity. \textbf{CoreSet} \cite{sener2017active} applies K-center selection for feature coverage. \textbf{DeepFool} \cite{ducoffe2018adversarial} targets points near decision boundaries via adversarial perturbations. \textbf{BALD} \cite{gal2017deep} estimates informativeness through Bayesian mutual information. \textbf{BADGE} \cite{ash2019deep} combines uncertainty and diversity using gradient embeddings. \textbf{BMMAL} \cite{shen2023towards} integrates multimodal informativeness and diversity for modality balance. All baselines share the same model backbones and training setup for fair comparison.

\subsection{Multimodal Active Learning Performance}
Table~\ref{tab:top1_results} reports top-1 accuracy with a fixed labelling budget of \textbf{3,000 samples}. 
As the training-set sizes differ substantially, this corresponds to different labelled fractions across datasets (6.6\% for Food101, 20.4\% for KineticsSound, and 2.9\% for VGGSound). 
RL\mbox{-}MBA achieves \textbf{0.8650} on Food101, \textbf{0.4841} on KineticsSound, and \textbf{0.2223} on VGGSound, outperforming all baselines at the same labelled-set size. 
Compared with the strongest baseline BMMAL (0.8609/0.4745/0.2053), RL\mbox{-}MBA yields consistent gains, with a particularly notable improvement on VGGSound, indicating better utilisation of multimodal complementarity under limited annotations.

Figure~\ref{fig:performance_curve} shows learning curves as the budget increases. 
RL\mbox{-}MBA remains competitive across all stages and typically leads other methods, demonstrating favourable accuracy--budget trade-offs on diverse multimodal benchmarks.

\subsection{Modality Contribution Analysis}
\begin{figure}[t]
	\centering
	\includegraphics[width=7cm]{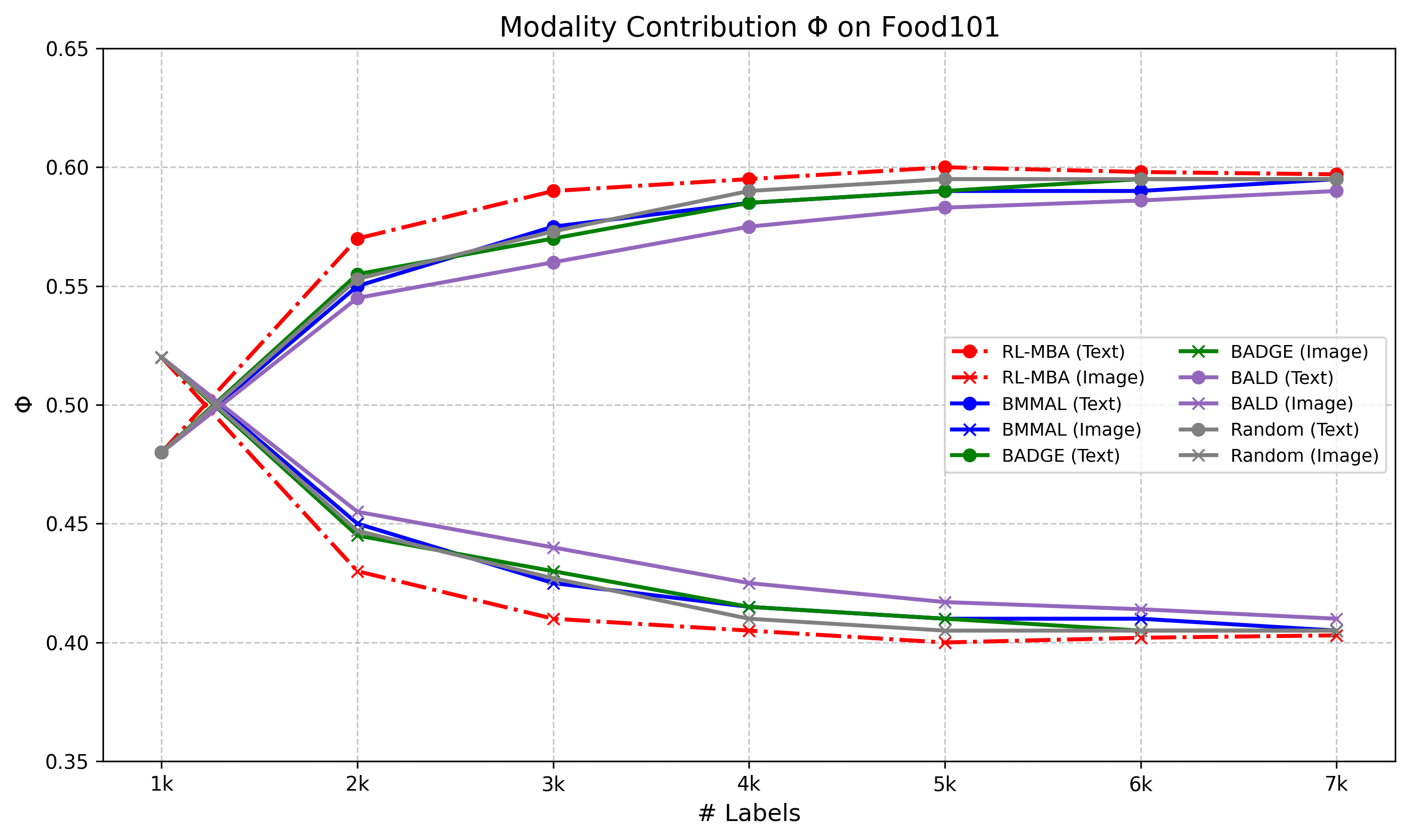}
	\caption{Modality contribution \(\phi\) on Food101 (1k–7k labels). 
		\(\phi\) is Shapley-derived; circles = Text, crosses = Image. 
		RL\mbox{-}MBA progressively increases text weight, while baselines remain static.}
	\label{fig:exp_food101_modality_contribution}
\end{figure}

To assess RL-MBA’s dynamic modality focus, we track the Shapley-derived contribution \(\phi\) across rounds. Figure~\ref{fig:exp_food101_modality_contribution} plots \(\phi\) on Food101 from 1{,}000 to 7{,}000 labels, comparing RL-MBA (red dash–dot) with BMMAL (blue), BADGE (green), BALD (purple), and Random (gray); circles denote text and “x” images. All methods start similarly, but RL-MBA gradually shifts toward text and de-emphasizes images, whereas baselines remain largely static. This stems from RL-MBA’s Adaptive Modality Contribution Balancing (AMCB) and Evidential Fusion for Difficulty-Aware Policy Adjustment (EFDA), which align sampling with the evolving value of each modality.

% 替换掉原来的两幅图：exp_ablation_kineticsound 与 exp_ablation_food101_reward_modes
\begin{figure}[t]
	\centering
	\includegraphics[width=0.87\linewidth]{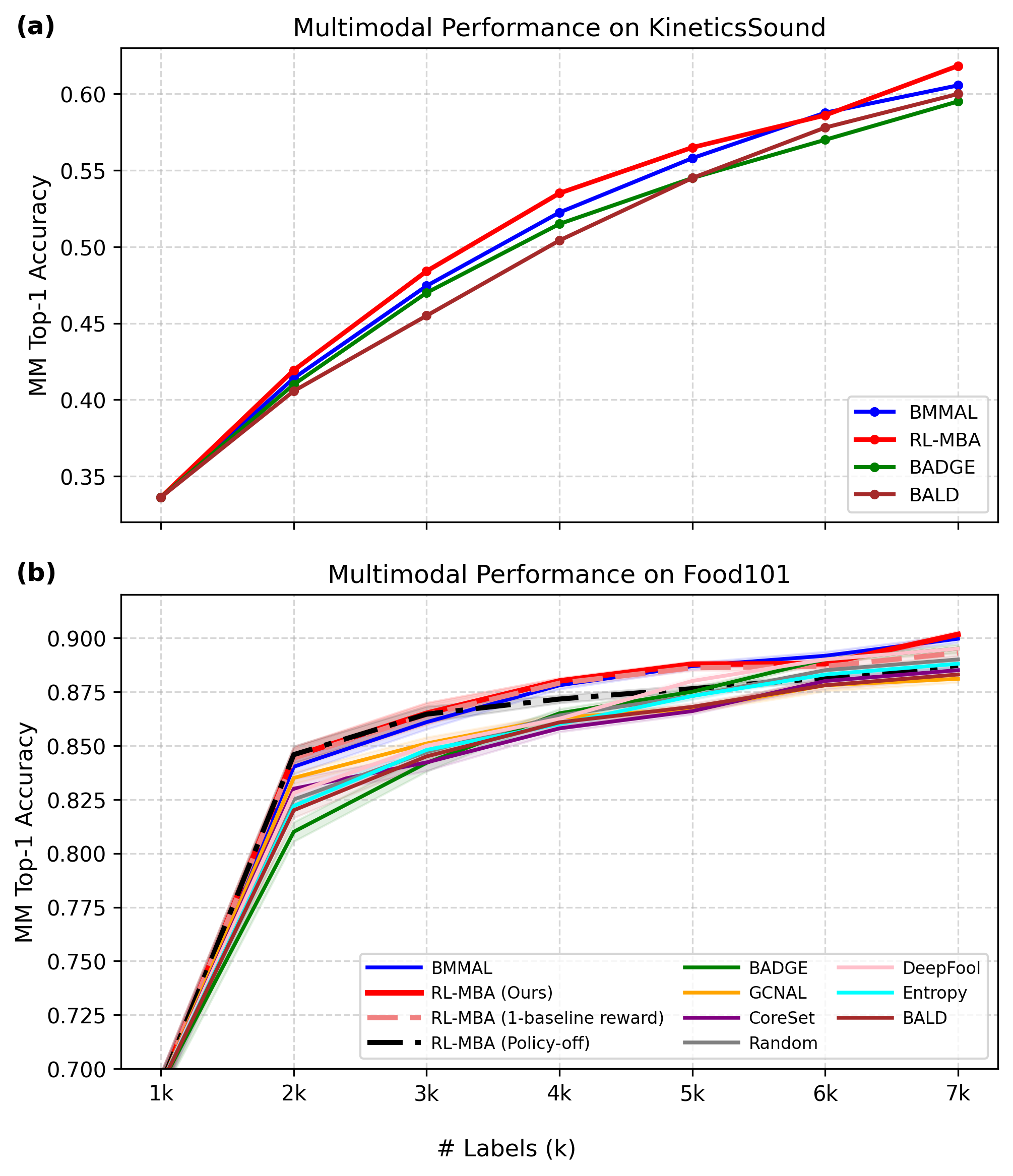} % 不写扩展名更稳
	\caption{Combined ablation and reward analysis. \textbf{(a)} Ablation on KineticsSound comparing BMMAL, RL-MBA w/ AMCB, RL-MBA w/ EFDA, and RL-MBA (Full). \textbf{(b)} Reward settings on Food101 comparing Relative, Absolute, and Incremental rewards.}
	\label{fig:exp_ablation_all}
\end{figure}

\begin{table}[t]
	\centering
	\fontsize{9}{12}\selectfont
	\setlength{\tabcolsep}{1.2mm}
	
	\begin{tabular}{lccc}
		\toprule
		\textbf{Variant} & \textbf{Food101} & \textbf{KineticsSound} & \textbf{VGGSound} \\
		\midrule
		BMMAL (Baseline) & 0.8609 & 0.4745 & 0.2053 \\
		RL-MBA w/ AMCB & 0.8621 & 0.4771 & 0.2059 \\
		RL-MBA w/ EFDA & 0.8637 & 0.4802 & 0.2177 \\
		RL-MBA (Full) & \textbf{0.8650} & \textbf{0.4841} & \textbf{0.2223} \\
		\bottomrule
	\end{tabular}
	
	\caption{Ablation study results (TOP-1 Accuracy) on Food101, KineticsSound, and VGGSound with 3,000 labels.}
	\label{tab:ablation}
\end{table}

\begin{table}[t]
	\centering
	\fontsize{7.6}{9}\selectfont
	\setlength{\tabcolsep}{1.2mm}
	\renewcommand{\arraystretch}{1.06}
	\caption{Food101 per-round wall-clock breakdown (s) with fixed Train/Val across methods. Round Total is computed as Pred$+$Sel$+$Train$+$Val$+$Policy.}
	\vspace{-6pt}
	\label{tab:food101_wallclock_norm}
	\begin{tabular}{lcccccc}
		\toprule
		Strategy & Pred & Sel & Train & Val & Policy Update & Round Total \\
		\midrule
		GCNAL		  & 450.15 & 700.54  & 639.16 & 56.11 & 0.00 & 1845.96 \\
		DeepFool	  & 263.12 & 1273.25 & 639.16 & 56.11 & 0.00 & 2231.64 \\
		BALD 	  	  & 519.21 & 0.44    & 639.16 & 56.11 & 0.00 & 1214.92 \\
		BADGE         & 162.10 & 312.84  & 639.16 & 56.11 & 0.00 & 1170.21 \\
		BMMAL         & 146.00 & 310.12  & 639.16 & 56.11 & 0.00 & 1151.39 \\
		RL\mbox{-}MBA & 155.41 & 33.48   & 639.16 & 56.11 & 0.23 & 884.39 \\
		\bottomrule
	\end{tabular}
\end{table}

\subsection{Efficiency}
Table~\ref{tab:food101_wallclock_norm} shows that RL\mbox{-}MBA achieves the lowest round total (884.39s), mainly by accelerating selection: \textbf{Sel} is 33.48s, compared with 312.84s (BADGE) and 310.12s (BMMAL). 
Meanwhile, the reinforcement-learning component incurs negligible overhead (\textbf{Policy Update} $=0.23$s), indicating that RL\mbox{-}MBA improves sampling efficiency without introducing substantial extra computation.

\subsection{Ablation Study}
We evaluate the contributions of RL-MBA’s components on Food101, KineticsSound, and VGGSound by comparing four settings: the BMMAL baseline, RL-MBA with only AMCB, RL-MBA with only EFDA, and the full method. Table~\ref{tab:ablation} (3{,}000 labels) further confirms this: on Food101, AMCB/EFDA bring incremental gains and RL-MBA reaches 0.8650; on KineticsSound and VGGSound, EFDA yields larger boosts, while RL-MBA attains the top scores (0.4841 and 0.2223), highlighting the benefit of jointly modeling modality balance and sample difficulty. On Food101, we also compare reward designs, Relative (Rel), Absolute (Abs), and Incremental (Delta) (Fig.~\ref{fig:exp_ablation_all}(b)). All start similarly at 1{,}000 labels, but as the budget grows the Relative reward stays highest (feedback-normalized and more adaptive).

\section{Conclusion}
We introduced RL-MBA, a reinforcement learning framework for multimodal active learning that adaptively balances modality contributions and prioritizes informative samples under limited labeling budgets. By coupling Adaptive Modality Contribution Balancing with Evidential Fusion for Difficulty-Aware adjustment, RL-MBA shows the benefit of reinforcement-driven adaptive sampling for mitigating modality imbalance and learning difficulty.

\small
\bibliographystyle{ieeenat_fullname}
\bibliography{main}

\end{document}